# Video Frame Interpolation via Structure-Motion based Iterative Fusion

Xi Li[1,2], Meng Cao[1], Yingying Tang[1], Scott Johnston[1], Zhendong Hong[1], Huimin Ma[2,3], Jiulong Shan[1]

Apple[1], Tsinghua University[2], University of Science and Technology Beijing[3]

## Abstract

*Video Frame Interpolation synthesizes non-existent images between adjacent frames, with the aim of providing a smooth and consistent visual experience. Two approaches for solving this challenging task are optical flow based and kernel-based methods. In existing works, optical flow based methods can provide accurate point-to-point motion description, however, they lack constraints on object structure. On the contrary, kernel-based methods focus on structural alignment, which relies on semantic and apparent features, but tends to blur results. Based on these observations, we propose a structure-motion based iterative fusion method. The framework is an end-to-end learnable structure with two stages. First, interpolated frames are synthesized by structure-based and motion-based learning branches respectively, then, an iterative refinement module is established via spatial and temporal feature integration. Inspired by the observation that audiences have different visual preferences on foreground and background objects, we for the first time propose to use saliency masks in the evaluation processes of the task of video frame interpolation. Experimental results on three typical benchmarks show that the proposed method achieves superior performance on all evaluation metrics over the state-of-the-art methods, even when our models are trained with only one-tenth of the data other methods use.*

## Author Keywords

Video frame interpolation; structure-motion information; iterative fusion; saliency based evaluation.

## 1. Introduction

Video Frame Interpolation (VFI) aims to provide a better visual experience [9, 2, 6]. This technology has a wide range of applications including frame rate conversion, adaptive animation rendering on high refresh rate devices, frame recovery in video streaming, and so on. In the past 10 years, deep learning technology has revolutionized many computer vision problems through spatial and temporal analysis on hierarchical features [12, 8, 9], which also provides inspiration for improving the performance of VFI.

Technically, VFI, given two or more adjacent frames, aims to generate a reasonable middle frame by learning the features of input images. When faced with this task, there are two key challenges. The first is to provide the correct occlusion relationship of moving objects, while the second is to maintain object structure. Figure 1 shows an example: In the middle frame, the motion of the balls should obey physical laws, as does the relative position between person's hand and other objects. To solve this problem, there are two approaches in existing works. The first is motion-based methods, which focus on the utilization of optical flow to provide an accurate point-to-point motion description [3, 2]. However, these approaches lack constraints on the imaging context. On the other hand, structure-based methods focus on object attribute alignment between the interpolated result and ground truth, which heavily relies on semantic and apparent features generated from hierarchical CNN layers [6, 8, 9]. This type of method provides results with better structural stability but tends to make the interpolated results blurred.

In this paper, we propose an end-to-end structure-motion based iterative fusion framework. Considering that the motion-based and structure-based methods have their own advantages, we first establish the two types of learning branches individually, followed by a unique fusion process utilizing adaptive attention masks. Both temporal and spatial information is considered by designing a temporal alignment unit for an enhancement between adjacent frames, and a spatial feature rectifier unit for optimization of hierarchical context. To further improve the fused results of feature representation from different source branches, an iterative process is adopted for complete integration. Based on the proposed iterative fusion framework, complementary feature representation from various sources is extracted and fused, therefore frame interpolation results of higher quality can be achieved.

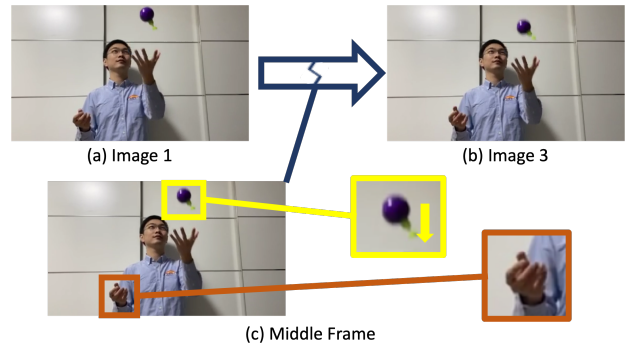

(a) Image 1     (b) Image 3

(c) Middle Frame

**Figure 1.** Example of video frame interpolation. In a generated middle frame, occlusion of the wall by the balls should obey physical laws, while the structure of the semantic regions (such as a hand) should be reasonable.

The main contributions of this paper are:

- First, a structure-motion based feature fusion method is proposed to exploit the advantages of different interpolated features via attention masks.

- Second, an iterative refinement framework is established to align subtle motion and structural changes from spatial and temporal information. Consequently, further improvement on accurate pointwise fusion is achieved.

- Third, results on three benchmarks demonstrate that our model achieves superior performance on all evaluation metrics over the state-of-the-art (SOTA) methods, using only one-tenth of the data used in training.

## 2. Related Works

*(a) Optical Flow based Frame Interpolation:* In existing CNN-based video frame interpolation, optical flow is widely used. Based on the temporal linear assumption for motion, a point-to-point mapping can be built from interpolated targets to the previous and next frames. End-to-end learning frameworks, for example ToFlow [16] and PWCNet [13], are popular CNN-based optical flow generation modules for general scenes. Specifically, PWCNet provides an efficient solution based on a coarse to fine process.

Based on motion estimation modules, a variety of methods are proposed to provide a better forward interpolation. MEMC-Net [3] proposes an occlusion mask to achieve an adaptive selection of interpolated results. Auxiliary information is utilized to improve interpolation performance. Xu *et al.* [15] provides a quadratic interpolation method to replace the linear assumption, while DAIN [2] introduced depth as an occlusion measure to weigh the contribution of different frames on different pixels. These methods provide reasonable interpolation and accurate point-to-point motion information. However, they are sensitive to the quality of estimated optical flow and lacks constraints on the object structure.

*(b) Kernel-based Frame Interpolation:* In contrast to methods based on optical flow, the kernel-based methods focus on multi-level feature alignment to achieve results with better maintenance of structure. In these methods, hierarchical low-level and high-level convolution features are encoded for synthetic results and ground truth. Using constraints from different semantic feature layers, the generated frames can achieve better object and scene structure stability and rationality.

To obtain structure based supervision, Niklaus *et al.* [9] provides a pretrained VGG network [11] to encode semantic features, while features from early layers in ResNet [7], as well as Laplacian feature and edges are used as a low-level context feature extractor [6]. In addition, deformable and dilated convolution [5, 17] modules are used in AdaCof [8] and FeatureFlow [6], to achieve larger and adaptive kernel receptive fields for better expression of different cases. In comparison, kernel-based learning pays more attention to multi-level information for structure stabilization. However, it may result blurred interpolation without point-to-point motion guidance.

Inspired by these works, we propose a structure-motion based iterative fusion framework, which uses learnable attention masks to exploit the advantages of different approaches. By fusing structure, motion, spatial and temporal information, better performance can be achieved with significantly less training data.

## 3. Overview

To achieve better video frame interpolation results, an end-to-end framework generates reasonable and clear intermediate frames. The proposed framework consists of two main components. Given two adjacent frames as inputs, first interpolated synthetic results based on structure and motion features are generated. Next, a post-processing learning structure using temporal-spatial information further optimizes the results. Figure 2 shows an overview of the proposed structure-motion based iterative fusion framework.

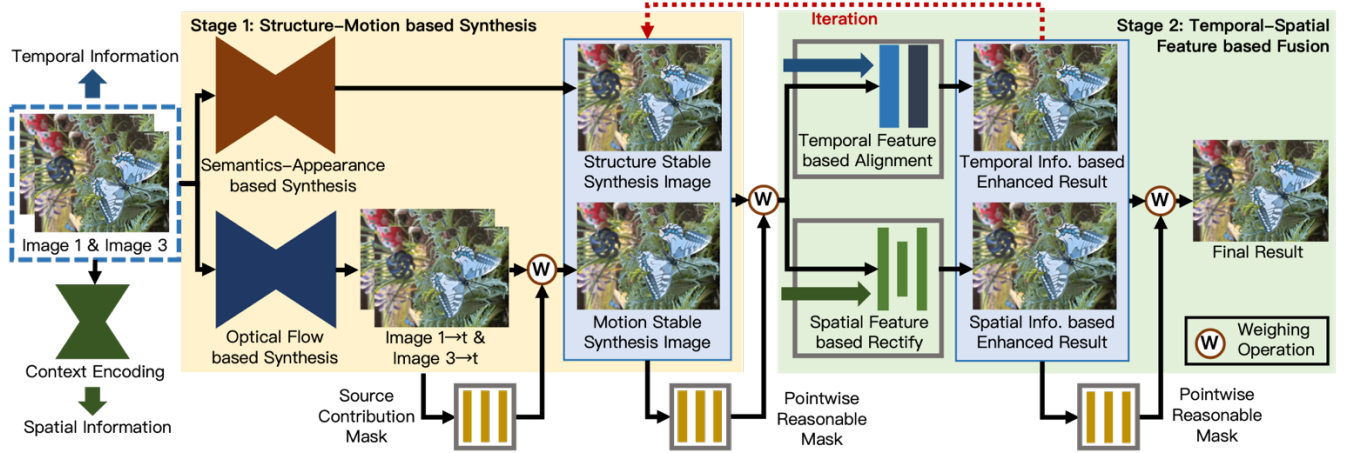

**Figure 2.** Overview of the proposed structure-motion based iterative fusion framework. In the first stage, structure stable and motion stable interpolation are generated respectively by semantics-appearance based module and optical flow based module. The two interpolated results are then fused by learnable weighing masks. In the second stage, a temporal feature based alignment unit and a spatial feature based rectifier unit are designed for result enhancement. The spatial and temporal information is then fully and precisely fused by a well-tuned iteration process, which results in superior interpolation quality.

## 4. Structure-Motion based Frame Synthesis

Given adjacent frames as input, two independent interpolation branches are introduced as guidance, which focus on structure rationality and motion accuracy respectively.

The first branch is a synthesis network learned from the semantic and apparent features. In this module, we use high-level and low-level features as supervision to achieve structure consistency between generated results and ground truth. Referring to [6], we introduce a learning framework built on the deformable convolutional operation to realize kernel-based frame interpolation. Next, convolution features and edge features are encoded to represent semantic and apparent information respectively. Feature alignment is learned at different levels between interpolated results and ground truth, to guide generation of structurally stable synthesized frames.

The second branch is a synthesis network learned from optical flow features. A pretrained PWCNet [13] is used for optical flow maps ($F_{1\to3}$, $F_{3\to1}$) calculation from input frames with bi-directional sequential order. For continuous frames, we assume that point-to-point motion relationships satisfy short-term linear constraint. Under this condition, the pixel mapping between the interpolated result and input frames is used to guide the forward interpolation process. The optical flow based synthesis module provides pointwise motion stable results.

The aforementioned two modules are pre-trained individually, followed by an end-to-end finetuning of complete VFI framework.

## 5. Iterative Enhancement for Feature Fusion

*(a) Mask based weighing:* To appropriately fuse interpolated results from different branches, we propose two types of adaptive attention masks: a source contribution mask and a pointwise attention mask.

First, in the motion-based synthesis branch, interpolated results generated from input images ($I_{1→t}$, $I_{3→t}$) can be obtained from optical flow maps ($F_{1→3}$, $F_{3→1}$) assuming short-term linear constraint. Meanwhile, two binary masks are calculated to annotate holes ($H_{1→t}$, $H_{3→t}$) during the forward frame interpolation. Since object or camera movement in adjacent frames may cause sudden changes of occlusion in different regions, a source contribution mask is learned to analyze different regions' dependency on both the previous and next frames, by taking two interpolated frames ($I_{1→t}$, $I_{3→t}$), optical flow maps ($F_{1→3}$, $F_{3→1}$) and the binary flow hole masks ($H_{1→t}$, $H_{3→t}$) as inputs.

Similarly, pointwise attention masks can be learned to generate the final interpolated frame from both structure-based and motion-based branches. Eq. (1) describes the attention mask weighing process. Given two synthesis results $I_{i_1}$ and $I_{i_2}$, $W(\cdot)$ is generated by the corresponding mask generation unit, the structure of which refers to [8]. Given the weighted summation process, the interpolated frame $I_o$ combines the advantages from individual branch at each image region,

$$I_o(\cdot) = W(\cdot) \cdot I_{in_1}(\cdot) + (1 - W(\cdot)) \cdot I_{in_2}(\cdot). \quad (1)$$

**(b) Spatial and Temporal based Enhancement:** To further improve the interpolation quality, we propose two refinement units to provide more precise adjustments on the attention mask fused interpolation results described in Section 5.1.

The first is a temporal feature based alignment unit. Inspired by [14] and [6], we observe that input frames and interpolated results usually have continuity in texture, edges, and other detailed attributes. The interpolated frame therefore can be further enhanced by using temporal correlated information. In our experiments, a variant of the network designed by the detailed enhancement module in [6] is adopted as the temporal feature alignment unit.

The second is a spatial feature based rectifier unit. For an interpolated frame, especially foreground regions with semantics, spatial context features can provide guidance to modify the representation of object structure and details. Therefore, given input frames, we first use hierarchical CNN features to represent context information. Then, similar to the post-processing modules in a variety of VFI works [2, 3], we also use a residual learning module to learn the difference between initially synthesized results and ground truth. That is to say, the summation of the initially interpolated frame and encoded residual can be represented as the output. In this rectifier unit, the inputs are the mask based fusion result and context features, and the output is the interpolated frame enhanced by spatial information.

## 6. Saliency based Evaluation

From the inference results encoded by the saliency calculation network, we observe that the masks can represent the tendency of people's attention when watching the video frames. Meanwhile, people have different quality requirements for foreground and background. Based on this, we define individual evaluation indicators for different regions generated by the saliency mask. Specifically, for each frame, we first build a saliency detection module to encode foreground and background masks ($M_f$ and $M_b$) for ground truth $I_{gt}$ in the evaluation process. In this paper, we use $U^2$-Net [10] to generate the saliency map, but this is replaceable for any saliency detection method. Further, for the generated result $I_{gen}$, the interpolation error $IE$ of foreground and background can be calculated respectively based on the masks, as Eq. (5),

$$IE_* = \frac{\sum_{(\cdot) \in I_{gt}} M_*(\cdot) \cdot |I_{gen}(\cdot) - I_{gt}(\cdot)|}{\sum_{(\cdot) \in I_{gt}} M_*(\cdot)}. \quad (5)$$

Here, ($\cdot$) represents a pixel, while $M_*(\cdot)$ is the corresponding saliency mask. The foreground and background interpolation error ($IE_f$ and $IE_b$) are weighted average absolute difference values.

Similarly, the mean square error (MSE) and peak signal to noise ratio (PSNR) for foreground and background can be calculated in the same way. The proposed indicators provide more reasonable metrics to evaluate interpolated frame quality for different regions, and also can be used as guidance for framework selection to produce tasks under disparate scenes.

## 7. Experiments

*(a) Dataset.* We use Vimeo-90K [16] as our training set. One thing to be noted, instead of using all 51,313 triplets, we build a subset of only 5,000 samples for model training. Less than 10% of the original data are selected to form our training set, as we observed a large number of redundant samples in the original dataset (similar image content and object motion rules). The reduced training dataset significantly improves training efficiency. More importantly, our models trained on this subset with the proposed end-to-end framework outperform state-of-the-art results trained on the full training dataset. The following three commonly used benchmarks are used in evaluation: Vimeo-90K *eval* set, UCF101 [12] and Middlebury [1]. The properties of these datasets are listed in Table 1.

**Table 1.** The properties of training and evaluation datasets.

| Dataset | Usage | Number of Samples | Image Sizes |
|---|---|---|---|
| Vimeo-90K (t) [16] | Training | 5000 | 448 × 256 |
| Vimeo-90K (e) [16] | Evaluation | 3782 | 448 × 256 |
| Middlebury [1] | Evaluation | 12 | Multi types |
| UCF-101 [12] | Evaluation | 379 | 256 × 256 |

*(b) Training Strategy.* We train the complete framework end-to-end. The mask based charbonnier penalty function $\rho(\cdot)$ [4] is used for the loss, and the constant value $c$ is set to $1e - 6$ in Eq.(6),

$$L = \rho(I_{gen}(x, y) - I_{gt}(x, y)). \quad (6)$$

*(c) Evaluated Metrics.* The interpolation error (IE), PSNR and Structural Similarity (SSIM) are used as indicators for objective quality evaluation. To align with human visual experience assessment, which naturally focuses more on the foreground, we do not only evaluate the indicators for the entire image, but also provide the foreground and background IE (F-IE, B-IE) and PSNR (F-PSNR, B-PSNR) based on saliency mask respectively.

*(d) Comparison with State-of-the-art Methods:* We compare our methods to state-of-the-art optical flow based and kernel-based video frame interpolation methods DAIN [2] and FeatureFlow [6] on three commonly used datasets. Table 2 illustrates the comparison details on objective evaluation indicators. The experimental results show that our framework achieves superior performance on all indicators over the state-of-the-art, even though our models are trained with only one-tenth of the data. The results also illustrate that all methods perform better on the background, as motion and/or structural changes are relatively moderate. This phenomenon not only shows the rationality of the proposed saliency-based evaluation, but also implies that the quality of the foreground region, which includes more complex and diverse motion changes, is the bottleneck for improving interpolated performance.

Table 2. Comparisons of methods on different datasets.

| Dataset | Method | PSNR (↑) | F-PSNR (↑) | B-PSNR (↑) | IE (↓) | F-IE (↓) | B-IE (↓) | SSIM (↑) |
|---|---|---|---|---|---|---|---|---|
| Vimeo-90K [16] | DAIN [2] | 34.96 | 32.98 | 37.14 | 2.21 | 3.30 | 1.81 | 0.9764 |
| | FeatureFlow [6] | 35.18 | 33.14 | 37.45 | 2.20 | 3.29 | 1.80 | 0.9760 |
| | **Ours** | **35.58** | **33.51** | **37.83** | **2.08** | **3.12** | **1.71** | **0.9788** |
| UCF-101 [12] | DAIN [2] | 35.00 | 31.98 | 38.16 | 2.81 | 4.91 | 2.15 | 0.9682 |
| | FeatureFlow [6] | 35.12 | 32.11 | 38.28 | 2.79 | 4.86 | 2.12 | 0.9689 |
| | **Ours** | **35.24** | **32.21** | **38.43** | **2.75** | **4.82** | **2.10** | **0.9692** |
| MiddleBury [1] | DAIN [2] | 36.82 | 34.90 | 38.72 | 2.04 | 2.82 | 1.70 | 0.9832 |
| | FeatureFlow [6] | 36.69 | 34.63 | 38.90 | 2.07 | 2.88 | 1.71 | 0.9835 |
| | **Ours** | **37.54** | **35.61** | **39.50** | **1.92** | **2.65** | **1.60** | **0.9858** |

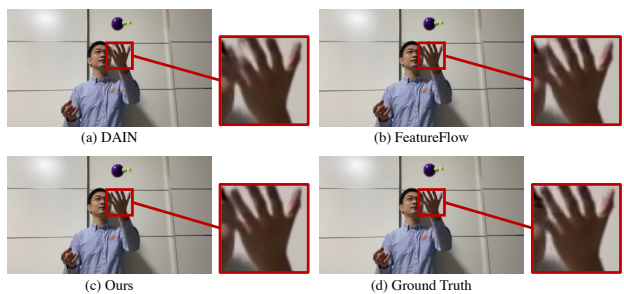

(a) DAIN  (b) FeatureFlow
(c) Ours  (d) Ground Truth

**Figure 3.** Qualitative results with VFI methods. Compare to frames generated via DAIN(a) and FeatureFlow (b), results provided by our methods (c) show more accurate and detailed expression. (d) shows ground truth.

Figure 3 illustrates several qualitative results. Compared to the SOTA methods, the result generated by our framework provide more clear expression, especially on detailed regions.

## 8. Conclusion

In this paper, we propose a novel video frame interpolation method via structure-motion based iterative fusion, which aims to provide results with a clear and reasonable appearance. To achieve this goal, a two-stage framework is established. Given two adjacent frames, we encode images by structure and motion based learning branches respectively in the first stage. Then, the temporal information alignment unit and spatial feature based rectifier unit is introduced in the second stage, which achieves further enhancement based on adjacent frames and hierarchical context. Here, iterative learning structure is utilized to integrate spatial and temporal feature based optimization, and hence to generate video results with higher quality. Experimental results on three commonly used benchmarks demonstrate that the proposed methods achieve superior performance on all evaluation metrics over the state-of-the-art, even though our models are trained with only one-tenth of the data.

## 9. Acknowledgements

We gratefully acknowledge the contribution form Dr. Wei Yao and Dr. Chaohao Wang for their suggestions and inputs to this work from display perspective.